\documentclass{isprs}
\usepackage{setspace}
\usepackage{geometry} 
\usepackage{makecell}
\usepackage{epstopdf}
\usepackage{tikz}
\usepackage{pgfplots}
\usepackage{amsmath}
\usepackage{amssymb}
\usepackage[caption=false]{subfig}

\begin{document}

\title{Bootstrapped CNNs for Building Segmentation on RGB-D Aerial Imagery}

\author{
 Clint Sebastian\textsuperscript{1}, Bas Boom\textsuperscript{2}, Thijs van Lankveld\textsuperscript{2}, Egor Bondarev\textsuperscript{1}, Peter H.N. De With\textsuperscript{1}}

 \address{
	\textsuperscript{1 }Eindhoven University of Technology, Eindhoven, The Netherlands - (c.sebastian, e.bondarev, p.h.n.de.with)@tue.nl\\
	\textsuperscript{2 }Cyclomedia B.V, Zaltbommel, The Netherlands - (bboom, tvanlankveld)@cyclomedia.com\\
 }


\commission{IV, }{IV} 
\workinggroup{IV/II} 
\icwg{}   

\abstract{
Detection of buildings and other objects from aerial images has various applications in urban planning and map making. Automated building detection from aerial imagery is a challenging task, as it is prone to varying lighting conditions, shadows and occlusions. Convolutional Neural Networks (CNNs) are robust against some of these variations, although they fail to distinguish easy and difficult examples.  We train a detection algorithm from RGB-D images to obtain a segmented mask by using the CNN architecture DenseNet. First, we improve the performance of the model by applying a statistical re-sampling technique called Bootstrapping and demonstrate that more informative examples are retained. Second, the proposed method outperforms the non-bootstrapped version by utilizing only one-sixth of the original training data and it obtains a precision-recall break-even of 95.10\% on our aerial imagery dataset. 
}


\maketitle


\section{INTRODUCTION}\label{INTRODUCTION}

\sloppy

Detection of terrestrial objects in aerial imagery is essential for applications such as urban planning, map making, change detection and disaster management. Due to structural data collection and database building by several institutes and companies, various government and private organizations are interested in maintaining a geospatial database of their municipalities. It is an expensive and time-consuming process for human experts to manually annotate each building in aerial imagery. Due to advancements in computer vision and machine learning along with recent improvements in hardware, it is now possible to automate various object recognition and detection tasks. Thus this paper aims at semantic segmentation of objects, where the objects are buildings or houses. However, building extraction from aerial imagery has a few inherent difficulties that make it a challenging problem. Varying lightning conditions, shadows and occlusions are common problems in aerial images. To alleviate the problems of shadows and different lighting conditions, we utilize a depth channel, which improves the robustness against these issues. Most of the older approaches to detect buildings from aerial images are based on a predefined set of features. However, recent works have employed learning-based approaches to detect buildings.

Although various machine learning algorithms are available, Deep Learning algorithms such as Convolutional Neural Networks (CNNs) have been proven effective for various image recognition and detection tasks. CNNs for building and road segmentation have been shown in \cite{mnih2010learning}, \cite{saito2015building}. Since its inception, it has become popular for various semantic object segmentation problems. Explosive growth of CNNs in the recent years has resulted in very deep architectures that offer state-of-the-art performance. While CNNs have made significant progress using complex architectures, it is still uncertain if a CNN exploits all the presented data to obtain the best results. As CNNs require a large amount of training data, it is imperative to choose the most informative training samples so as to obtain the best performance. However, feeding in specific samples that contribute to the improvement of the learning procedure of a CNN is not an easy task. For larger datasets, the challenge with the previous approaches is that they are not able to learn from the complex cases that occurs infrequently. As a consequence, an optimal performance is not achieved even if the algorithm is trained for a long period of time.

\begin{figure}[t]
\begin{center}
   \includegraphics[width=1.0\linewidth]{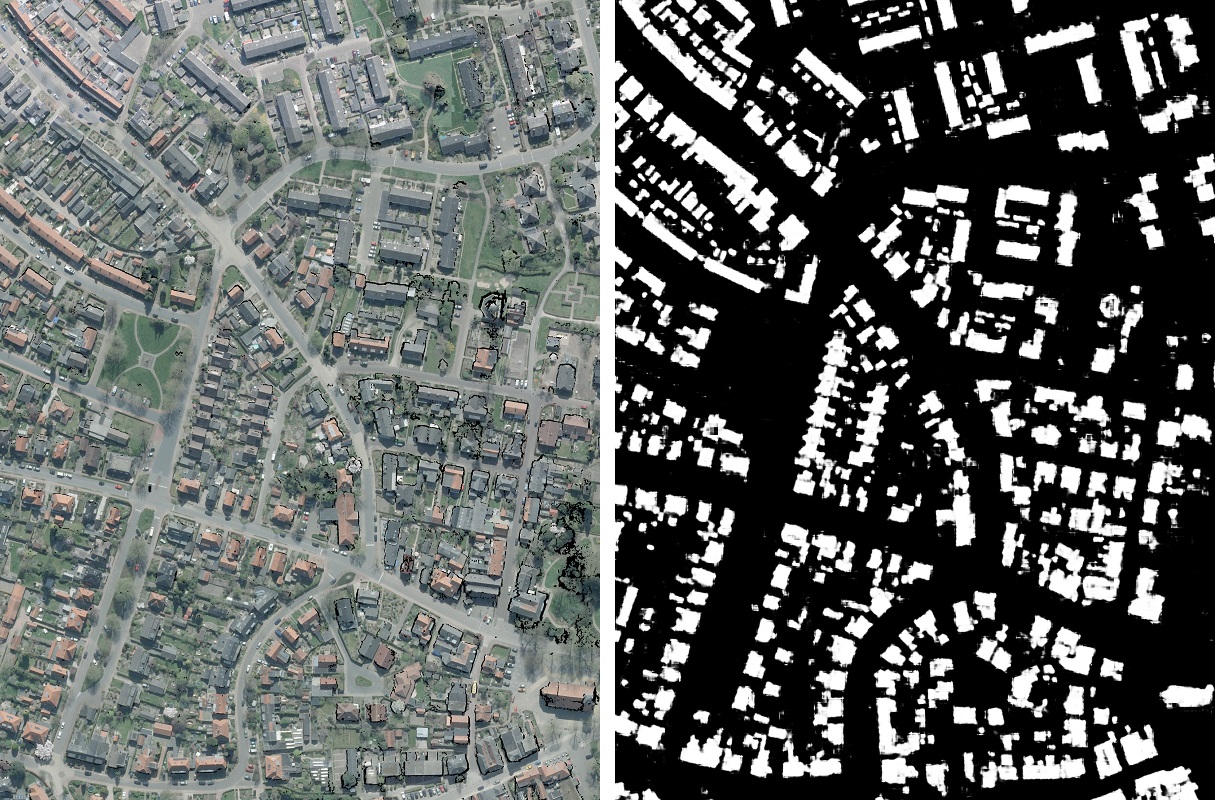}
\end{center}
   \caption{An aerial image (area of 370 $\times$ 495 meters) and its respective output produced by our method.}
\label{fig:intro}
\end{figure}

 In this research, we have developed a system based on a high-performance CNN architecture called Densely Connected Convolutional Network (DenseNet) for our building segmentation problem. Our first contribution is to improve the performance of the model, where we apply a statistical re-sampling technique called Bootstrapping \cite{efron1994introduction}. Through Bootstrapping, we establish a better balance between easy and difficult examples to generate a model that is suited for high-performance automated building detection. Unlike previous methods, we compute statistics across the entire training set and generate a new subset that retains the most important samples. Our second contribution is that the model trained on the subset not only outperforms the non-bootstrapped version, but needs to utilize only one-sixth of the original training data for retraining. We have also conducted experiments on further iterations of bootstrapping and additionally report these findings as well.

\section{RELATED WORK}\label{sec:RELATED WORK}

Satellite imagery has grown steadily over the past decades and a large amount of research has been conducted in extracting buildings and other objects from aerial imagery.
Rule-based classification of buildings and other terrestrial objects from LIDAR data was introduced by Forlani \textit{et al.} \cite{forlani2006complete}. Ali \"Ozg\"un Ok \cite{ok2008robust} proposed a rule-based method to detect buildings from aerial imagery. It involves a series of operations, featuring removal of irrelevant objects, multi-stage processing of edges and generation of vector polygons to detect buildings. A survey of different automated building detection techniques with depth data is conducted by Khoshelham \textit{et al.} \cite{frontoni2008comparative}. Dempster-Shafer and Adaboost methods have shown better overall performance and Bayesian methods are found to be effective in areas where buildings have similar heights. Sirma\c cek \textit{et al.} \cite{5523977} have proposed a method to detect buildings using structural features and probability theory. The structural features are extracted using a steerable filter set. This is tested on multiple aerial images taken from different sensors and is found to have high robustness.

An ensemble-learning approach for building detection is presented by Nguyen \textit{et al.} \cite{nguyenaerial}. Feature modalities are trained separately using Random Forest classifiers and the outputs are combined with Stacked Graphical Models. Final inference of the object is obtained from the fusion of multiple discriminative probabilistic classifiers. Inference of roads from aerial images by exploiting Neural Networks was introduced by Mnih and Hinton \cite{mnih2010learning}. An extension of this work to detect roads and buildings simultaneously was presented by Saito and Aoki \cite{saito2015building}. They trained a CNN with image patches and the resultant output is a multi-channel segmented output. Multiple pathways in CNNs for encompassing both global and local context has shown to improve performance by \cite{DBLP:journals/corr/Marcu16}
    
Recent growth of research efforts in CNNs has led to a multitude of CNN architectures. Example such as GoogLeNet, ResNet and DenseNet have proven their effectiveness in various image recognition challenges. GoogLeNet \cite{Szegedy_2015_CVPR} features a 22-layer deep architecture with inception modules: a component which processes the same input with multiple convolution filters. ResNet or Deep Residual Network \cite{he2015deep} introduce identity connection to fit the underlying mapping and avoid the problem of vanishing gradient by an ensemble of shallower networks \cite{veit2016residual}. A similar approach is taken by DenseNet \cite{huang2016densely}, where the mapping is received from all the preceding layers. The obtained feature maps are concatenated into a single tensor instead of the identity mapping in ResNets, where the connections are combined via summation. This concatenation allows better feature re-usability and receives partial supervision from the earlier layers with shorter connections. 
 
For many learning algorithms, class imbalance is a common issue that reduces the quality of results. To mitigate this problem, many learning algorithms employ Bootstrapping (also known as hard-negative mining) to strengthen the classifier \cite{felzenszwalb2010object}, \cite{dollar2009integral}, \cite {rowley1998neural}. Some of the recent work has applied hard negative mining in CNNs. Shrivastava \textit{et al.}  \cite{shrivastava2016training} proposed to incorporate hard Regions of Interests (RoI) for the region proposals in their object detection framework. An online bootstrapping approach for segmentation is proposed by Wu \textit{et al.} \cite{DBLP:journals/corr/WuSH16a}, where the hard training pixels are taken from a given mini-batch rather than the whole training set. Our method is similar to this approach, however, we compute the loss of all the training samples to obtain a global behavior to generate a set of bootstrap samples.

\section{Preliminary Experiments}

\subsection{Dataset}

The dataset consists of five municipalities in The Netherlands, namely Arnhem, Eindhoven, Hellevoetsluis, Heerenveen and Zutphen. Each municipality covers an area of roughly 5 km$^2$ with a pixel resolution of 0.01~m$^2$ per pixel. The dataset consists of the RGB images along with an additional depth channel. Each depth pixel represents a vertical distance to the corresponding estimated ground level. The dataset is divided into training, validation and test sets
in approximately the ratio 70:10:20, respectively, for all experiments. Due to the large resolution of the aerial images (5000 $\times$ 5000 pixels), each image is divided into patches. We empirically determined that inclusion of context is beneficial. Therefore, we used an input patch size of 80 $\times$ 80  (RGB-D image) for a 24 $\times$ 24 pixel segmented output.  Combination of local and global contexts has been shown to improve performance. In \cite{DBLP:journals/corr/Marcu16}, two networks are jointly trained such that they predict a label patch. However, considering the pixel resolution of our dataset, this is a computationally expensive task. Therefore, we select only a single context as input for training our network. 

\begin{figure}[h]
\begin{center}
   \includegraphics[width=1.0\linewidth]{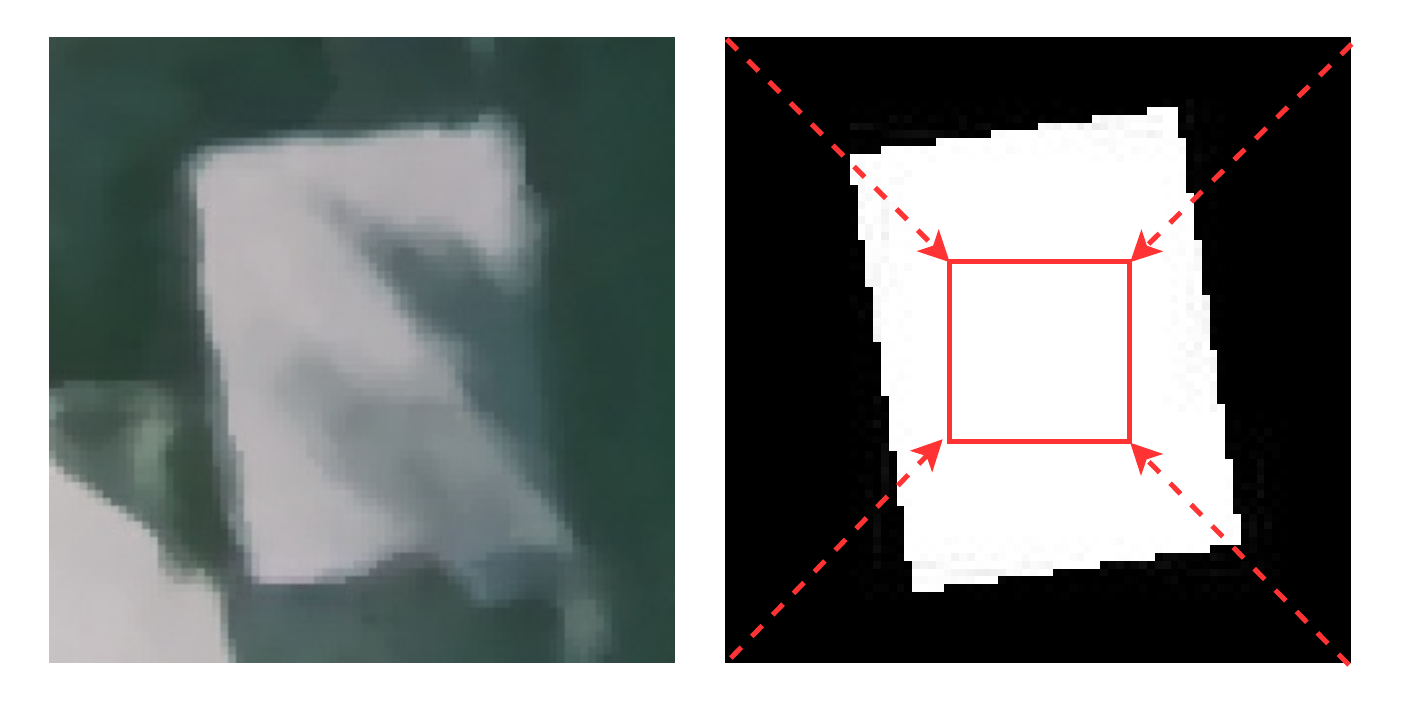}
\end{center}
   \caption{RGB patch of an aerial image of size 80 $\times$ 80 pixels and its corresponding ground truth. The area of 24 $\times$ 24 pixels inside the ground truth is the target of the network.}
\label{fig:long}
\end{figure}

\begin{figure*}[t]
\begin{center}
   \includegraphics[width=1.00\linewidth]{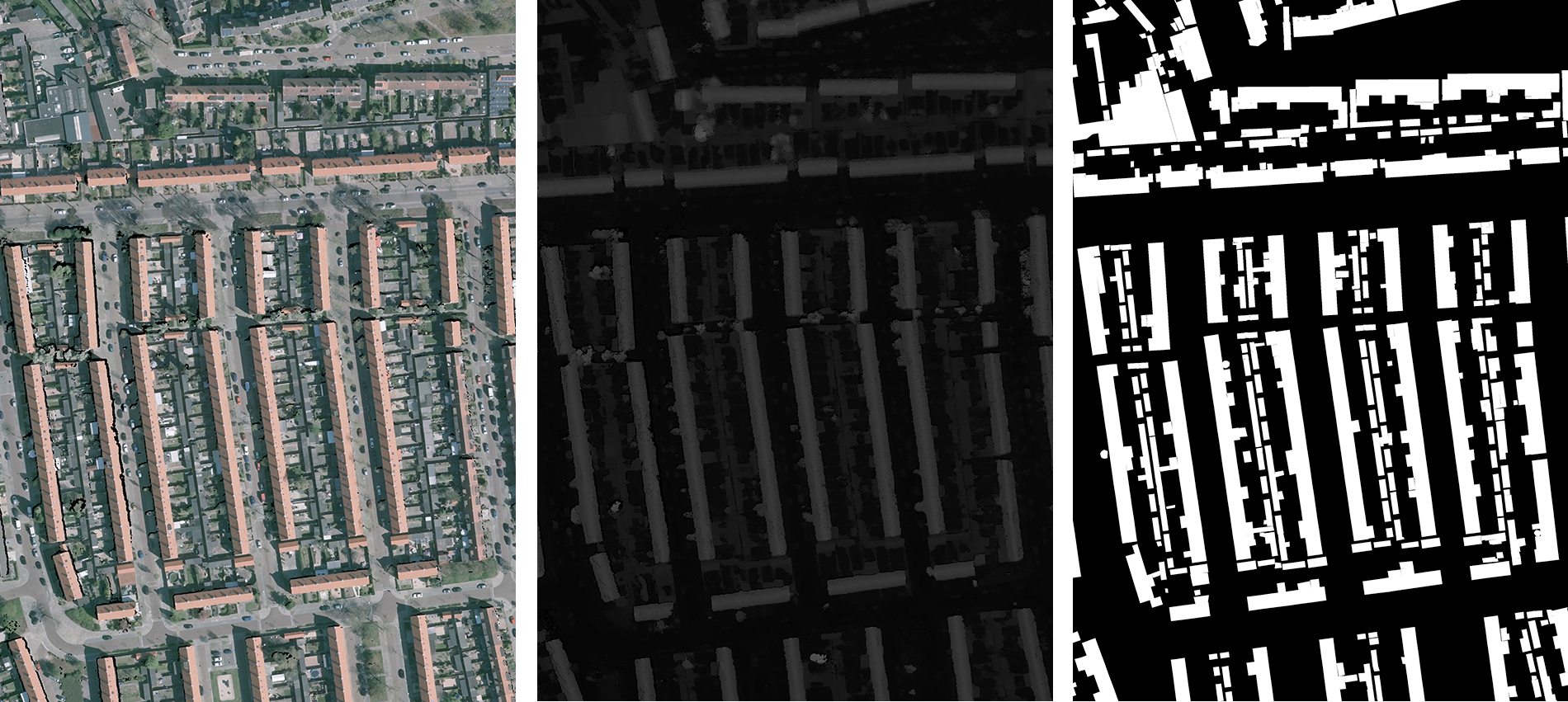}
\end{center}
   \subfloat[\label{RGB}RGB image]{\hspace{.33\linewidth}}
   \subfloat[\label{height} Depth map]{\hspace{.33\linewidth}}
   \subfloat[\label{gt} Ground truth.]{\hspace{.33\linewidth}}
   \caption{ Example of an RGB-D input image (area of 250 $\times$ 340 meters) and its respective groundtruth. }
\label{fig:dataset}
\end{figure*} 

\subsection{Evaluation Metric and building overlap}
The output produced by the CNN is evaluated by the number of successfully segmented buildings. The common metric used for evaluation is the precision-recall curve. At building level evaluation, \emph{recall} implies the number of buildings detected out of the total number of buildings whereas \emph{precision} is the number of buildings detected from the total number of segmented buildings. The quality of the results is analyzed using the precision-recall break-even point, i.e. the point where the precision and recall values are equal. Intuitively, this is when the number of missing buildings are equal to the number of incorrect segmented outputs. Segmentation of buildings with perfect overlap is difficult to achieve. Even a single missing pixel over the prediction heavily penalizes the results. This leads to inaccurate quantification of the results. Hence, an additional parameter called ``overlap" is introduced. 

The overlap parameter is the number of building pixels in the output that is present with respect to its ground-truth pixels. For example, an overlap of 50\% implies that 50\% of the building pixels are correctly segmented with respect to its ground truth. For the experiments in this research, we measure the precision-recall curve at different overlaps to give a broader perspective of the performance.

\subsection{Implementation details and preliminary results}
The preliminary experiments are performed using the AlexNet architecture. However, we remove the last fully connected layer and reshape the output to an \textit{n} $\times$ \textit{n} patch.
We also perform a K-fold analysis by removing one municipality for testing, while the model was trained with the other four. We observed consistent performance in the K-fold analysis, that ensures the reliability and generalization of the trained model. This results in a break-even
point of  78.69~$\pm$~0.26\% at 90\% building overlap with the ground truth and 92.82~$\pm$~0.15\% break-even at 50\% building overlap.

\subsection{DenseNet for Building Segmentation}
\begin{figure}[htp!]
\begin{center}
   \includegraphics[width=1.0\linewidth]{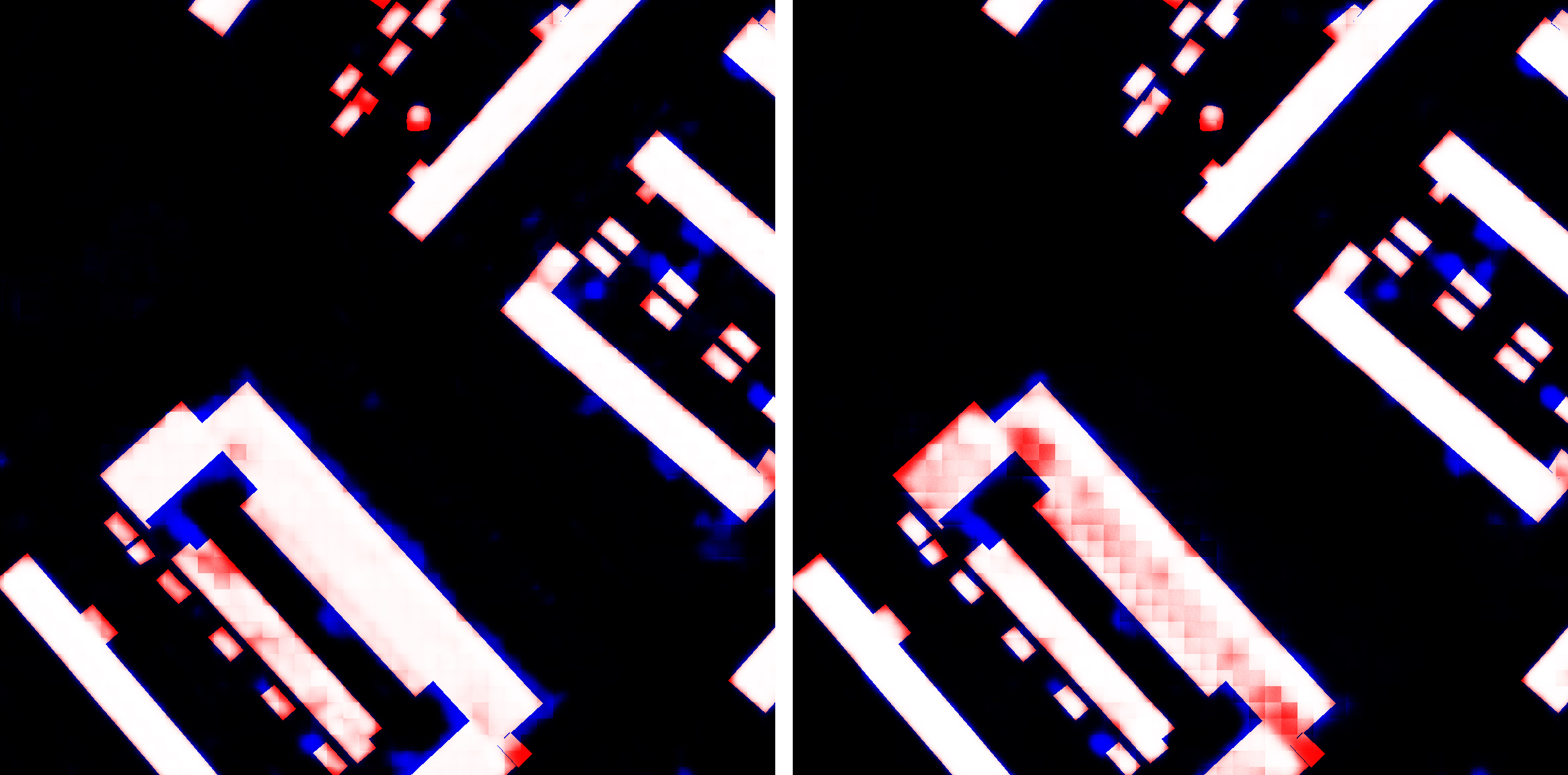}
\end{center}
   \caption{ Results from AlexNet (left) and DenseNet (right). White and black colors indicate true positives and true negatives. Red and blue colors indicate false negatives and false positives. AlexNet classifies pixels more accurately than DenseNet in a few cases.}
\label{fig:alex_vs}
\end{figure}

\begin{figure*}[t]
\begin{center}
   \includegraphics[width=1.00\linewidth]{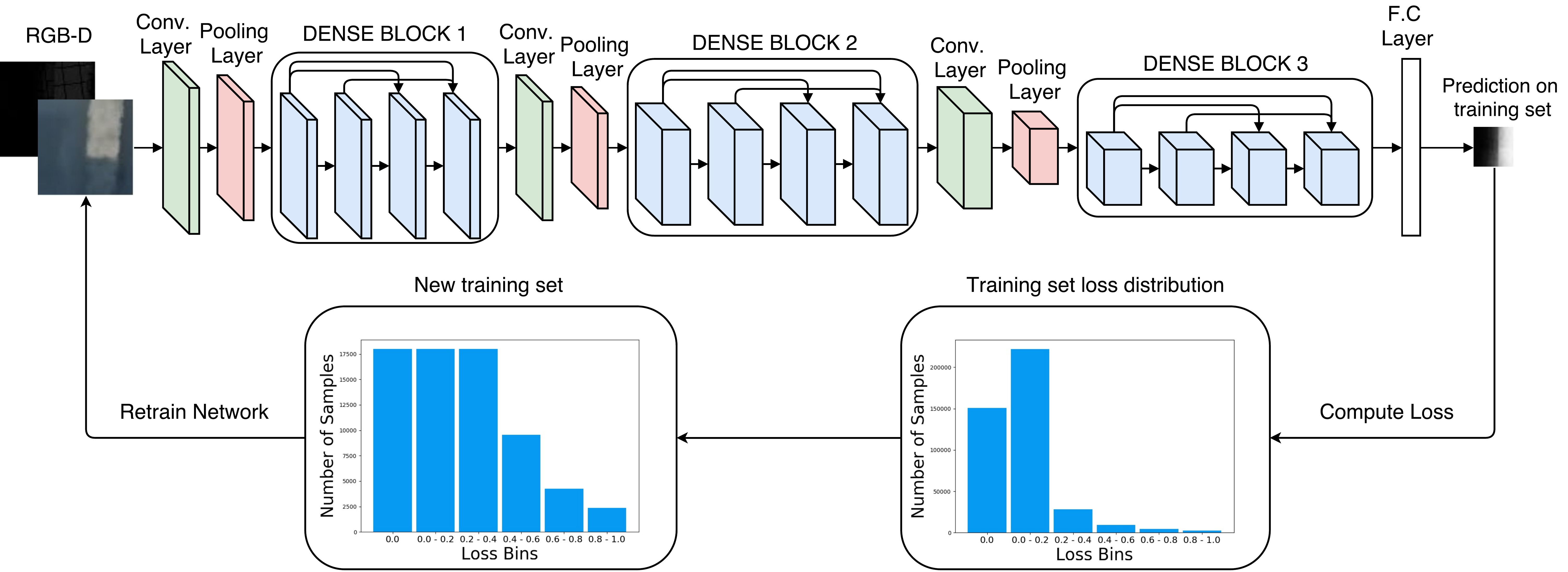}
\end{center}
   \caption{ Bootstrapping framework with DenseNet. An RGB-D patch is the input for the DenseNet. The trained model is used for inference on the training set and the training samples are re-sampled based on loss distribution to retrain the network.}
\label{fig:archit}
\end{figure*}

Recent works have introduced architectures that do not follow  strict sequential pathways \cite{he2015deep}  \cite{huang2016densely}. An architecture that utilizes such a topology is the DenseNet (see Figure~\ref{fig:archit}), for which every layer is connected to all lower layers. This type of connectivity enables feature re-usability and has high parameter efficiency. The resulting connectivity for a layer $\textbf{x}_{\ell}$ of this network is defined as: 
\begin{equation}
\textbf{x}_{\ell} = F_{\ell}([\textbf{x}_{\ell - 1}, \textbf{x}_{\ell - 2}, \dot{....}, \textbf{x}_{0}]),
\end{equation}
where [...] is the concatenation operation and $F(\cdot)$ is defined as a composition of Batch Normalization, Rectified Linear Unit (ReLU), convolution and dropout \cite{ioffe2015batch} \cite{srivastava2014dropout} \cite{maas2013rectifier}.
An attractive property of this network during backpropagation is that each layer receives weight updates from all its higher layers. Updating the weight of the layer $\textbf{x}_{\ell}$ in a Dense block is:
\begin{equation}
\frac{\partial\mathcal{L}}{\partial \textbf{x}_{\ell}} = 
\frac{\partial\mathcal{L}}{\partial \textbf{x}_{k}}
\Big(\sum_{i=\ell + 2}^{k} 
\Big(\prod_{d=i}^{k} \Big(\frac{\partial \textbf{x}_{d}}{\partial \textbf{x}_{d - 1}} \Big)\Big)
\cdot 
\frac{\partial \textbf{x}_{i - 1}}{\partial \textbf{x}_{\ell}}
\Big),
\end{equation}

where $k$ is the growth of the dense block and $\mathcal{L}$ the error obtained from the loss function. For our segmentation problem, we modify the DenseNet by adding a max-pooling layer after the first convolution layer and replacing the global pooling layer with a max-pooling layer. We add two fully connected layers, where the last one is reshaped into a 24 $\times$ 24~pixel patch. We use three dense blocks and set the growth rate of the network to 12. In addition, we have applied dropout with a ratio of 0.1 to all the convolution layers in a dense block. With DenseNet, the precision-recall break-even point improves to 94.78~\% at 50~\% building overlap compared to 92.82~\% of AlexNet. Although this network gives better results, an interesting observation is that these networks behave differently on the same training data. For some building pixels, AlexNet classifies more accurately than DenseNet. However, we did not observe any particular type of buildings having better segmentation with DenseNet. We further investigate this phenomenon in the following section.

\section{Bootstrapping CNNs}

We observe that both networks learn in a slightly different way from the same data, see Fig.~\ref{fig:alex_vs}. Instead of training a network for long periods of time, which may not improve the performance, we focus on improving by presenting more informative samples. Typically, the training samples contributing to the learning process have equal priority in the mini-batches. Instead, we provide the samples in a controlled fashion. The naive approach is to re-train only on the difficult samples, but as CNNs are highly sensitive to the input data, it would start to overfit on the difficult samples. To achieve a robust learning, we add a limited number of easy samples for bootstrapping, in addition to the difficult samples from the training set.
 
Let \big\{$(x_n, y_n), n = 1,...,N$\big\} represent the training data, where $x_n$ and $y_n$ are the input sample (RGB-D image) and its ground truth,  respectively. We describe our model as
$\hat{y_n} = \varphi(x_n, \mathcal{D})$, where $\hat{y_n}$ is the predicted output of the input sample $x_n$ with a model $\varphi$ trained over the data $\mathcal{D}$. We minimize the loss $\mathcal{L}(y_n, \hat{y_n})$ on our training set until it reaches a minimum on the validation set.  The cross-entropy loss $\mathcal{L}(y_n, \hat{y_n}) \in [0,1]$ of a prediction $\hat{y_n}$ with respect to its ground truth $y_n$ is defined as: 
\begin{equation}
\mathcal{L}(y_n, \hat{y_n}) = -\dfrac{1}{P} \sum\limits_{i=1}^{P} \Big[ p_i\cdot \log(\hat{p_i}) + (1 - p_i)\cdot \log(1 -\hat{p_i}) \Big] ,
\end{equation}
where $p_i$ is the probability of the pixel in the ground truth $y_n$ and $\hat{p_i}$ is the probability of the same pixel in the prediction $\hat{y_n}$. We classify each $x_n$ in the training set into bins $(a,b]$ with respect to its loss $\mathcal{L}(y_n, \hat{y_n})$. An additional bin for samples with a zero cross-entropy loss is also maintained. Instead of training the model on the complete dataset 
$\mathcal{D}$, we train on subset $\mathcal{D}_k \subset \mathcal{D}$, such that the $\sum_{n=1}^{N} \mathcal{L}(y_n, \hat{y_n})$ obtained from $\varphi(x_n, \mathcal{D}_k)$, has lower validation error than that of the model $\varphi(x_n, \mathcal{D})$. This bootstrapping scheme is shown in Fig. \ref{fig:archit}.

Initially, the DenseNet is trained with the original dataset, which consists of patches. Each training sample is tested using the trained model and the cross-entropy loss is computed. After computing the loss, each sample is assigned to a bin with a given step size. The samples that have a perfect prediction (cross-entropy loss of zero) are assigned to a separate bin. We have conducted our experiments with a bin step size of 0.2. A step size of 0.2 results in six loss bins [0], (0,~0.2], (0.2,~0.4], (0.4,~0.6], (0.6,~0.8] and (0.8,~1.0]. We observe that most trained samples have a low loss: almost 90\% of this samples have a cross-entropy loss less than 0.2\%. To incorporate the difficult examples, we collect all the samples that have a cross-entropy loss larger than 0.2. However, training only on the difficult samples can result in overfitting. It is often the case that easy-to-learn samples that have redundant features and occur frequently, mask the presence of difficult samples during training. Therefore, along with difficult examples, we add easy examples. We select a random subset of easy examples (cross-entropy loss between 0 and 0.2) with an equal size as hard examples (cross-entropy loss between 0.2 - 1.0). We train a DenseNet model from scratch using the new subset of data comprising of the balanced easy and difficult examples. 

\begin{table}[h]
\begin{center}\small
\begin{tabular}{|c|c|c|c|}
\hline
Overlap & AlexNet & DenseNet & Bootstrapping \\
\hline\hline
25\% \ & - & 4,033 (95.95) & 4,045 (96.24)\\
\hline
50\% \ & 3,901 (92.82) & 3,984 (94.78) & 3,997 (95.10)\\
\hline
75\% \ & - & 3,852 (91.64) & 3,886 (92.45)\\
\hline
90\% \ &  3,307 (78.69) & 3,529 (83.98) & 3,644 (86.77)\\
\hline
\end{tabular}
\end{center}
\caption{Number of detected buildings and PR break-even point at different building~overlaps on the test set. The last column denotes improved results after bootstrapping DenseNet.}
\end{table}
\begin{figure*}[tbhp!]

\begin{center}
   \includegraphics[trim={0 17.3cm 0 0},clip,width=1.0\linewidth]{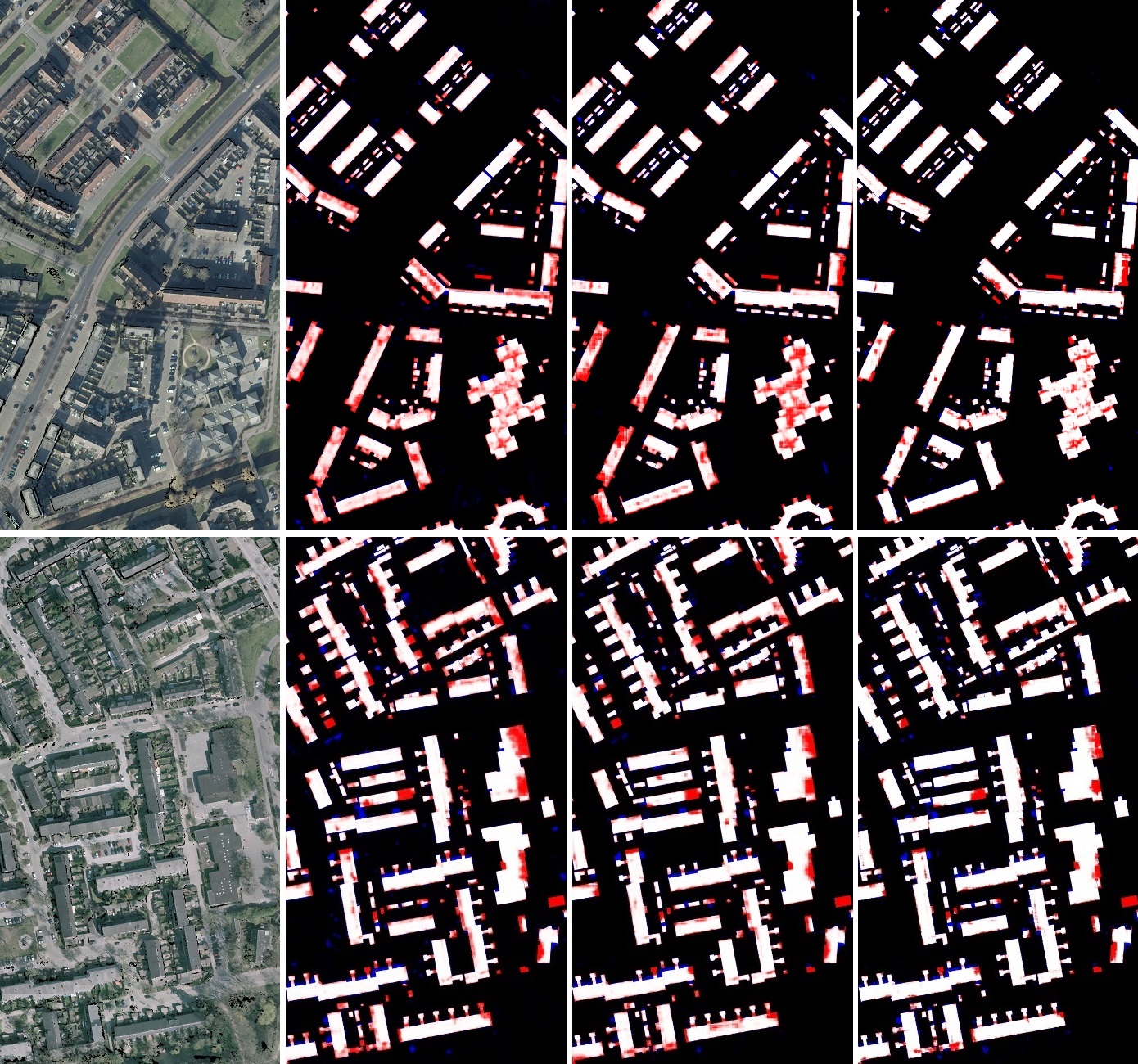}
\end{center}
   \subfloat[\label{RGB_image} RGB image]{\hspace{.25\linewidth}}
   \subfloat[\label{AlexNet} AlexNet]{\hspace{.25\linewidth}}
   \subfloat[\label{DenseNet} DenseNet]{\hspace{.25\linewidth}}
   \subfloat[\label{DenseBoot} DenseNet + Bootstrapping]{\hspace{.25\linewidth}}

   \caption{ RGB image (area of 213 $\times$ 399 meters) and outputs from different networks. All networks are trained with RGB-D inputs (depth not shown). Large buildings segmented by AlexNet show better performance compared to DenseNet. However, more buildings are detected by DenseNet. After bootstrapping, DenseNet detects more buildings and improves the confidence of existing predictions.}
\label{fig:final_comp}
\end{figure*}
After a round of bootstrapping, the performance improves on the test set. Since we select random samples from the zero and (0,~0.2] bins, we conduct the same experiment multiple times and consistently observe improved performance. After bootstrapping, we obtain a break-even point of 95.10 $\pm$ 0.14 at an overlap of 50\% and a break-even point of 86.77 $\pm$ 0.19 at an overlap of 90\%. We find that out of a total of 4,203 buildings on the test set, 12 extra buildings are found because of the bootstrapping at an overlap of 25\%. As the overlap ratio is increased, the performance gap widens, resulting in a gain of 115 buildings at an overlap of 90\%. Through bootstrapping, we improve not only the detection rate, but also obtain a higher confidence of the existing detections. Results are shown in Table 1 and Fig. \ref{fig:final_comp}.
\section{Further rounds of Bootstrapping}
We continue to generate bootstrapped training sets until the training loss converges to a minimum. However, the loss does not decrease on the test set and instead gives an oscillating performance at each bootstrapping round (see Table 2). To understand this phenomenon, we average the loss for the same samples that contributed towards the improved learning in the first bootstrapping iteration. For example, if 1,000 samples were present in the bin (0.8,~1.0] after the original training iteration, we compute the average loss for the same samples in all other bootstrapping iterations. Note that these samples may be present in a different loss bin at a different bootstrap iteration.

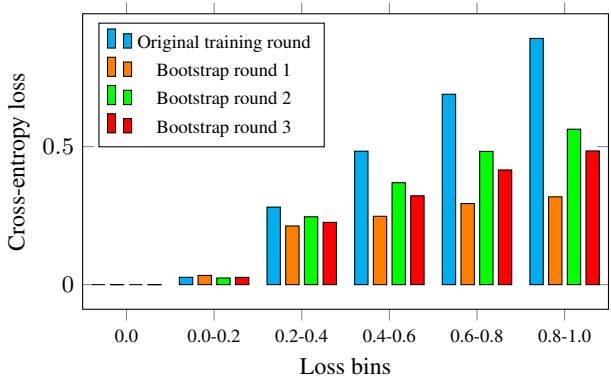
\begin{figure}[h]
\begin{center}
\begin{tikzpicture}
  \begin{axis}[
  	legend columns=4,
    transpose legend,
    legend entries={{\small Original training round},{\small Bootstrap round 1},{\small Bootstrap round 2},{\small Bootstrap round 3}}
  	x label style={at={(axis description cs:0.5,0.0)},anchor=north},
    y label style={at={(axis description cs:0.02,0.5)},anchor=north},
    xlabel=Loss bins,
    ylabel=Cross-entropy loss,
    anchor=south,
    width=8.5cm,
    height=5.5cm,
    ybar,
    enlarge x limits=0.1,
    legend style={at={(0.5,0.5)},
    anchor=north,legend columns=-1},
    bar width=5pt,
   symbolic x coords={0.0, 0.0-0.2, 0.2-0.4, 0.4-0.6, 0.6-0.8, 0.8-1.0},
    xtick=data,
    x tick label style={rotate=0,anchor=north, font=\scriptsize},
    legend pos=north west
    ]
    \addplot [ybar,fill=cyan] coordinates {
    (0.0, 0.0)
    (0.0-0.2, 0.0267)
    (0.2-0.4, 0.2809)
    (0.4-0.6, 0.4835)
    (0.6-0.8, 0.6898)
    (0.8-1.0, 0.8922)
    };
    \addlegendentry{{\scriptsize Original training round}}
    \addplot [ybar,fill=orange] coordinates {
    (0.0, 0.0001)
    (0.0-0.2, 0.0334)
    (0.2-0.4, 0.2126)
    (0.4-0.6, 0.2477)
    (0.6-0.8, 0.2938)
    (0.8-1.0, 0.3182)
    };
    \addlegendentry{{\scriptsize Bootstrap round 1}}
    \addplot [ybar,fill=green] coordinates {
    (0.0, 0.0)
    (0.0-0.2, 0.0241)
    (0.2-0.4, 0.2458)
    (0.4-0.6, 0.3694)
    (0.6-0.8, 0.4828)
    (0.8-1.0, 0.5632)
    };
    \addlegendentry{{\scriptsize Bootstrap round 2}}
    \addplot [ybar,fill=red] coordinates {
    (0.0, 0.0001)
    (0.0-0.2, 0.0263)
    (0.2-0.4, 0.2257)
    (0.4-0.6, 0.3219)
    (0.6-0.8, 0.4155)
    (0.8-1.0, 0.4844)
    };
    \addlegendentry{{\scriptsize Bootstrap round 3}}

  \end{axis}
\end{tikzpicture}
\caption{Cross-entropy on the training set at different bootstrap iterations of DenseNet. Samples at each bootstrap round are sorted by their loss in the order of the original training round. }
\label{fig:plot}
\end{center}
\end{figure}

From Figure \ref{fig:plot}, we observe the oscillating behavior of the training set. This oscillatory behavior is in agreement with varying break-even points in the test set (Table 2) at different bootstrap iterations (from loss bins (0.2-0.4] and higher). This phenomenon happens because the model overfits on the training set and starts to swap the easy and difficult examples at each iteration of the bootstrapping. When the actual difficult examples are used for training during bootstrapping, the performance reverts to the improved state. This case is observed for the first time after sampling from initial training and noticeable at the second iteration of Bootstrapping. Therefore, the difficult samples obtained after original training control the performance on the test set. 
\begin{table}[htp!]
\begin{center}\small
\begin{tabular}{|c|c|c|c|c|}
\hline
Training rounds & Original & 1 &  2 & 3 \\
\hline
\thead{Cross-Entropy loss \\ (training set) } & 0.0569 & 0.04109 & 0.04078 & 0.03914 \\
\hline
\thead{PR Break-even \\ (test set) } & 0.8398 & 0.8677 &0.7539 &0.8593 \\
\hline

\end{tabular}
\end{center}

\caption{Average cross entropy loss of the training samples at each training round of DenseNet. For bootstrap training rounds (denoted as 1,2 and 3), the training loss decreases. However, the precision-recall break-even point of the corresponding test set oscillates.}
\end{table}

\section{Conclusions}

We have proposed a framework that segments RGB-D aerial images through Bootstrapping and compared the performance of AlexNet and DenseNet. We observe that AlexNet outperforms DenseNet in certain regions and propose to improve the efficiency of the DenseNet by giving priority to the difficult examples. We introduce a method of bootstrapping that considers the whole training set to generate bootstrap samples. Using the newly generated bootstrap training set, we have trained a model from scratch, utilizing only one-sixth of the training data. As a result, the bootstrapped model outperforms the CNN at all overlap levels and improves the performance by a significant margin at larger overlaps. Finally, we have studied further iterations of bootstrapping and found that the examples that are selected as difficult by the original training round, are essential for performance improvement.

{
	\begin{spacing}{0.9}
		\bibliography{ISPRSguidelines_authors} 

\begin{thebibliography}{xx}

\bibitem[Dollar et al., 2009]{dollar2009integral}
Dollar, P., Tu, Z., Perona, P. and Belongie, S., 2009.
\newblock Integral channel features.
\newblock In: \emph{Proc. BMVC}, pp.~91.1--91.11.
\newblock doi:10.5244/C.23.91.

\bibitem[Efron and Tibshirani, 1994]{efron1994introduction}
Efron, B. and Tibshirani, R.~J., 1994.
\newblock {\em An introduction to the bootstrap}.
\newblock CRC press.

\bibitem[Felzenszwalb et al., 2010]{felzenszwalb2010object}
Felzenszwalb, P.~F., Girshick, R.~B., McAllester, D. and Ramanan, D., 2010.
\newblock Object detection with discriminatively trained part-based models.
\newblock {\em IEEE transactions on pattern analysis and machine intelligence}
  32(9), pp.~1627--1645.

\bibitem[Forlani et al., 2006]{forlani2006complete}
Forlani, G., Nardinocchi, C., Scaioni, M. and Zingaretti, P., 2006.
\newblock Complete classification of raw lidar data and 3d reconstruction of
  buildings.
\newblock {\em Pattern Analysis and Applications} 8(4), pp.~357--374.

\bibitem[Frontoni et al., 2008]{frontoni2008comparative}
Frontoni, E., Khoshelham, K., Nardinocchi, C., Nedkov, S. and Zingaretti, P.,
  2008.
\newblock Comparative analysis of automatic approaches to building detection
  from multi-source aerial data.
\newblock In: \emph{Proceedings GEOBIA 2008-Pixels, Objects, Intelligence
  GEOgraphic Object Based Image Analysis for the 21st Century, Calgary, Canada,
  5-8 August 2008; IAPRS, XXXVIII (4/C1), 2008}, International Society of
  Photogrammetry and Remote Sensing (ISPRS).

\bibitem[He et al., 2016]{he2015deep}
He, K., Zhang, X., Ren, S. and Sun, J., 2016.
\newblock Deep residual learning for image recognition.
\newblock In: \emph{Proceedings of the IEEE conference on computer vision and
  pattern recognition}, pp.~770--778.

\bibitem[Huang et al., 2017]{huang2016densely}
Huang, G., Liu, Z., van~der Maaten, L. and Weinberger, K.~Q., 2017.
\newblock Densely connected convolutional networks.
\newblock In: \emph{The IEEE Conference on Computer Vision and Pattern
  Recognition (CVPR)}.

\bibitem[Ioffe and Szegedy, 2015]{ioffe2015batch}
Ioffe, S. and Szegedy, C., 2015.
\newblock Batch normalization: Accelerating deep network training by reducing
  internal covariate shift.
\newblock In: \emph{Proceedings of the 32nd International Conference on
  International Conference on Machine Learning - Volume 37}, ICML'15, JMLR.org,
  pp.~448--456.

\bibitem[Maas et al., 2013]{maas2013rectifier}
Maas, A.~L., Hannun, A.~Y. and Ng, A.~Y., 2013.
\newblock Rectifier nonlinearities improve neural network acoustic models.
\newblock In: \emph{Proc. ICML}, Vol.~30number~1.

\bibitem[Marcu, 2016]{DBLP:journals/corr/Marcu16}
Marcu, A., 2016.
\newblock A local-global approach to semantic segmentation in aerial images.
\newblock {\em Computing Research Repository}.

\bibitem[Mnih and Hinton, 2010]{mnih2010learning}
Mnih, V. and Hinton, G.~E., 2010.
\newblock Learning to detect roads in high-resolution aerial images.
\newblock In: \emph{European Conference on Computer Vision}, Springer,
  pp.~210--223.

\bibitem[Nguyen et al., n.d.]{nguyenaerial}
Nguyen, T., Kluckner, S., Bischof, H. and Leberl, F., n.d.
\newblock Aerial photo building classification by stacking appearance and
  elevation measurements.

\bibitem[Ok, 2008]{ok2008robust}
Ok, A., 2008.
\newblock Robust detection of buildings from a single color aerial image.
\newblock {\em Proceedings of GEOBIA 2008} p.~6.

\bibitem[Rowley et al., 1998]{rowley1998neural}
Rowley, H.~A., Baluja, S. and Kanade, T., 1998.
\newblock Neural network-based face detection.
\newblock {\em IEEE Transactions on pattern analysis and machine intelligence}
  20(1), pp.~23--38.

\bibitem[Saito and Aoki, 2015]{saito2015building}
Saito, S. and Aoki, Y., 2015.
\newblock Building and road detection from large aerial imagery.
\newblock In: \emph{SPIE/IS\&T Electronic Imaging}, International Society for
  Optics and Photonics, pp.~94050K--94050K.

\bibitem[Shrivastava et al., 2016]{shrivastava2016training}
Shrivastava, A., Gupta, A. and Girshick, R., 2016.
\newblock Training region-based object detectors with online hard example
  mining.
\newblock In: \emph{Proceedings of the IEEE Conference on Computer Vision and
  Pattern Recognition}, pp.~761--769.

\bibitem[Sirmacek and Unsalan, 2011]{5523977}
Sirmacek, B. and Unsalan, C., 2011.
\newblock A probabilistic framework to detect buildings in aerial and satellite
  images.
\newblock {\em IEEE Transactions on Geoscience and Remote Sensing} 49(1),
  pp.~211--221.

\bibitem[Srivastava et al., 2014]{srivastava2014dropout}
Srivastava, N., Hinton, G.~E., Krizhevsky, A., Sutskever, I. and Salakhutdinov,
  R., 2014.
\newblock Dropout: a simple way to prevent neural networks from overfitting.
\newblock {\em Journal of Machine Learning Research} 15(1), pp.~1929--1958.

\bibitem[Szegedy et al., 2015]{Szegedy_2015_CVPR}
Szegedy, C., Liu, W., Jia, Y., Sermanet, P., Reed, S., Anguelov, D., Erhan, D.,
  Vanhoucke, V. and Rabinovich, A., 2015.
\newblock Going deeper with convolutions.
\newblock In: \emph{The IEEE Conference on Computer Vision and Pattern
  Recognition (CVPR)}.

\bibitem[Veit et al., 2016]{veit2016residual}
Veit, A., Wilber, M. and Belongie, S., 2016.
\newblock Residual networks are exponential ensembles of relatively shallow
  networks.
\newblock {\em arXiv preprint arXiv:1605.06431}.

\bibitem[Wu et al., 2016]{DBLP:journals/corr/WuSH16a}
Wu, Z., Shen, C. and van~den Hengel, A., 2016.
\newblock High-performance semantic segmentation using very deep fully
  convolutional networks.
\newblock {\em Computing Research Repository}.

\end{thebibliography}
	\end{spacing}
}

\end{document}